# Vision Transformer with Convolutions Architecture Search

Haichao Zhang, Kuangrong Hao, Witold Pedrycz, *Life Fellow, IEEE*, Lei Gao, *Member, IEEE*, Xuesong Tang, and Bing Wei, *Member, IEEE*

*Abstract*—Transformers exhibit great advantages in handling computer vision tasks. They model image classification tasks by utilizing a multi-head attention mechanism to process a series of patches consisting of split images. However, for complex tasks, Transformer in computer vision not only requires inheriting a bit of dynamic attention and global context, but also needs to introduce features concerning noise reduction, shifting, and scaling invariance of objects. Therefore, here we take a step forward to study the structural characteristics of Transformer and convolution and propose an architecture search method—Vision Transformer with Convolutions Architecture Search (VTCAS). The high-performance backbone network searched by VTCAS introduces the desirable features of convolutional neural networks into the Transformer architecture while maintaining the benefits of the multi-head attention mechanism. The searched block-based backbone network can extract feature maps at different scales. These features are compatible with a wider range of visual tasks, such as image classification (32 M parameters, 82.0% Top-1 accuracy on ImageNet-1K) and object detection (50.4% mAP on COCO2017). The proposed topology based on the multi-head attention mechanism and CNN adaptively associates relational features of pixels with multi-scale features of objects. It enhances the robustness of the neural network for object recognition, especially in the low illumination indoor scene.

## I. INTRODUCTION

Convolutional neural networks (CNNs) have been dominating the area of neural networks for approximately a decade, with succesful architectures reported in ImageNet [1] image classification challenges, such as AlexNet [2], VGGNet [3], and ResNet [4]. Correspondingly, CNN gradually becomes more and more powerful through adopting more parameters and more diverse forms of convolutional architecture. As the backbone network for visual tasks, CNN has achieved significant performance improvements by adopting a variety of network architectures and received extensive attention in the field of computer vision. On the other hand, Transformer [5, 6] which occupies a dominant position in natural language processing (NLP) [7], has recently started to show its role in computer vision. Vision Transformer (ViT) [8] is a computer vision model that completely relies on the Transformer architecture to achieve competitive performance on large-scale datasets. ViT [8] makes the smallest modifications to the Transformer [5, 6] architecture, enabling its adaption from natural language understanding to computer vision tasks. It splits one image into discrete non-overlapping patches, which can be regarded as tokens in NLP. In this way, ViT [8] expresses the rough location space information as a special location code and models global relations in the standard Transformer architecture to realize image classification.

Although ViT has achieved success on multiple datasets, it lacks robustness for object recognition in complex scenes. One possible reason is that the multi-headed attention mechanism (MHSA) in ViT [8] focuses more on the relationship between pixels, which makes it lack highly desirable features as shifting, and scaling invariance of objects. For example, when CNN is oriented to images with local structure, it can utilize the local receptive field, shared weights, and spatial subsampling to achieve a strong degree of shifting and scaling invariance. In addition, the hierarchical learning pattern of the convolution kernel considers the complex local spatial background from simple low-level edges and textures to high-order semantic patterns. These features are not available in ViT [8]. It is worthy of exploring a way to integrate CNN features with MHSA features.

Neural architecture search (NAS) [9-11] provides an effective solution to the above-mentioned problems. In most cases, CNN needs to be designed according to specific tasks, which leads to endless hyperparameter tuning. NAS has been proved to be a very promising method to alleviate the tedious work, and the process of designing the neural network architecture has been taken as an optimization problem. So far, there are mainly three approaches for NAS. One of them is the evolutionary algorithm [11]. This approach generates new individuals through evolutionary operations and obtains the best individuals by searching in the population. Another approach is reinforcement learning, which utilizes a recursive network as the controller and generates the network

This work was supported in part by the Fundamental Research Funds for the Central Universities (2232021A-10, 2232021D-37), National Natural Science Foundation of China (61903078), Natural Science Foundation of Shanghai (20ZR1400400, 21ZR1401700), and the Fundamental Research Funds for the Central Universities and Graduate Student Innovation Fund of Donghua University (CUSF-DH-D-2021048).

Haichao Zhang, Kuangrong Hao, Xuesong Tang, Bing Wei are with the College of Information Science and Technology, Donghua University, and also with the Engineering Research Center of Digitized Textile and Apparel Technology, Ministry of Education, Donghua University, Shanghai 201620, China. (e-mail: dhu_zhc@mail.dhu.edu.cn; krhao@dhu.edu.cn, *Corresponding author*; tangxs@dhu.edu.cn; bingwei@dhu.edu.cn)

W. Pedrycz is with the Department of Electrical and Computer Engineering, University of Alberta, Edmonton, AB T6R 2V4, Canada, and also with the Department of Electrical and Computer Engineering Faculty of Engineering, King Abdulaziz University, Jeddah 21589, Saudi Arabia, and also with the Systems Research Institute, Polish Academy of Sciences, 01-447 Warsaw, Poland (e-mail: wpedrycz@ualberta.ca).

L. Gao is now with CSIRO, Waite Campus, Urrbrae, SA 5064, Australia.(e-mail: Lei.Gao@csiro.au)

architecture through reinforcement learning. Although the performance of the above two approaches is impressive, for large datasets such as ImageNet, their search processes are highly resource-intensive. Therefore, the search approach on the gradient-based NAS has received widespread attention because it greatly speeds up the search process and achieves impressive performance. For this reason, Differentiable Architecture Search (DARTS) [12-14] is proposed resulting in outstanding performance. By relaxing the discrete search space to its continuous version, DARTS [13] describes the process of searching for the optimal neural network architecture as a two-layer optimization method and develops an optimized architecture relative to the performance of the verification set through the gradient descent method. Because the gradient-based method shows better performance on ImageNet, the gradient-based idea is adopted to fuse CNN features and MHSA features.

In response to the above clues and ideas, the Vision Transformer with Convolution Architecture Search (VTCAS) algorithm is proposed, and an optimal block topology is formed to combine convolution (Conv) operations with MHSA operations. On this basis, a backbone network with multi-scale output is obtained which obtains excellent classification performance. Under the model parameters of 31M parameters, it reaches the classification performance of Top-1 82% on ImageNet. At the same time, this network has achieved outstanding performance for object detection tasks as the backbone. In some low illumination scenes, the VTCAS-based object detection framework shows stronger recognition ability. While constructing the network, we also solve the problem of the disappearance of the gradient caused by the data transformation of the feature map between the MHSA operation and the Conv operation. The algorithm successfully fuses MHSA and CNN features and comes with the following advantages:

1) A Vision Transformer with Convolutions Architecture Search (VTCAS) algorithm is proposed, which optimizes the correlation between the relational features of pixels and the multi-scale feature of objects by a gradient-based method. The proposed method enhances the robustness of the neural network for object recognition in the conditions of low illumination present in indoor scenes.

2) The proposed topology involving MHSA and CNN enables neural architecture search on the ImageNet dataset directly. At the same time, the gradient discontinuity problem caused by the feature map transition between the MHSA operation and the Conv operation is solved effectively.

3) The network obtained by VTCAS search achieves excellent performance in image classification and object detection. Especially in low illumination scenes, it has a strong ability to identify objects.

The paper is organized as follows. Section II introduces and summarizes related studies. In Section III, the VTCAS with significantly improved feature extraction capability is provided. The experiments design in Section IV and the searched network that show a strong ability to identify objects in low illumination scenes are discussed in Section V. Finally, we conclude with a summary of our findings and comment on possible future directions in Section VI.

## II. LITERATURE REVIEW

CNN has played a crucial role in the field of computer vision and evolved into a standard network model. CNN has been around for decades, but it is not until the introduction of AlexNet [2] and VGG [3] that they become a dominant model for computational vision tasks. Since then, deeper and more efficient convolutional network architectures have been proposed to further advance the development of computer vision, such as ResNet [4], DenseNet [15], and HRNet[16]. In addition, previous work on individual variants of convolution such as depthwise separable convolution, deformable convolution, and ECANet [17] have been proposed and widely promoted. Although CNN and its variants are widely used as the backbone in computer vision, Transformer-like network architectures show potential for unified modeling between vision and language. In Transformer-like network architectures, MHSA can effectively extract relationships among "tokens" in a sequence and analyze objects with complex structures. In this paper, this method is integrated into a CNN with features of local receptive fields, shared weights, and spatial subsampling. In the following, by Vision Transformer and NAS, we will represent the background of the proposed approach.

*A. Vision Transformer*

Vision Transformer (ViT) [8] is the first architecture to demonstrate that the Transformer architecture can achieve better performance on large-scale datasets as ImageNet [1]. Specifically, ViT splits each image into sequential tokens to form patches and utilizes multiple Transformer layers consisting of MHSA and Positionwise Feed-forward module (FNN) layers to these patches. Similarly, DeiT [18] further explores efficient training methods and network distillation for ViT. The Swin Transformer [19] architecture has the flexibility to model at different scales and a linear computational complexity relative to the image size. These characteristics make it compatible with a wide range of vision tasks, including image classification and object detection.

To better adapt to the classification task, some works have introduced design alteration. Correspondingly, the Conditional Position encoder Vision Transformer [20] replaces the pre-defined embedding position in ViT, which enables the Transformer block to process input images of any size without interpolation. Transformer-iN-Transformer [21] utilizes external Transformer blocks to deal with the embedding of the patch and the relationship between the pixels, respectively. Tokens-to-Token (T2T) [22] improves ViT mainly by connecting multiple tokens in a sliding window to one token. This operation is very different from the downsampling of the convolutional network, especially in the details of the normalization, the connection of multiple tokens has greatly increased the complexity of computing. Pyramid vision transformer (PVT) [23] adopts a multi-scale and multi-stage architecture design method to obtain good performance on intensive prediction tasks. Our proposed algorithm fully considers the advantages of MHSA and combines the scaling invariance of CNN to achieve target recognition in complex scenes.

*B. Architecture Search with Block*

In recent years, neural architecture search receives widespread attention, and numerous algorithms have been developed. Evolutionary algorithms [24, 25] have been applied to search for high-performance architectures, such as evolutionary network topologies. With the continuous deepening of research, the methods based on evolutionary computation continue to show satisfactory performance [26-29]. Similarly, NAS based on reinforcement learning also shows excellent performance. One can refer here to the pioneering work of applying the RNN network as a controller sequential design architecture [30]. The gradient-based NAS method [12-14] utilizes the gradient of network architecture parameters to optimize the architecture. This method greatly speeds up the search process to obtain a well-performing network architecture.

Deep neural networks have complex discrete architecture and continuous hyperparameters, so searching the architecture of such neural networks is a very challenging task. Therefore, many researchers [31] have simplified the process of NAS to search the shared blocks, and these shared blocks finally constitute the target neural network. Initially, NASNet [32] searches for shared blocks on smaller datasets and then transfers these shared blocks to larger datasets. The shared block idea proposed by NASNet provides a new method for NAS. Later, Sun et al. [33] proposed a neural architecture search algorithm based on a genetic algorithm to search the shared block. Similarly, the idea of block architecture search is also used in the reinforcement learning NAS. BlockQNN [34] proposes to use a proxy network to generate the shared blocks. Most gradient-based NAS methods utilize block optimization for network architecture search. DARTS [13] proposes an architecture search algorithm for optimizing shared blocks in proxy networks. P-DARTS [12] solves the problem of the distance between the target network and the proxy network in DARTS through a progressive search method. In our work, inspired by the basic ideas in P-DARTS, a new block topology is proposed to progressively search for the optimal network architecture by combining desirable features of MHSA operation and Conv operation.

## III. PROPOSED METHOD

In this section, the details of the proposed framework of VTCAS are described. We start with the overview of VTCAS to present the process of the proposed approach on proxy networks. Then, the search space and evaluation network architecture are introduced in detail. Last, the feature map transformation process to solve gradient disappearance is described in detail.

*A. Algorithm Overview*

In this work, we leverage P-DARTS and Swin Transformer [19] as the baseline framework. The goal of our method is to search the topology of a block with Conv operations and MHSA operations, which can construct a network of $L$ blocks to identify complex objects. Taking into account the complex structural characteristics of Conv operations and MHSA operations, the proposed algorithm models this task as a bilevel optimization problem, which can be expressed as:

$$\min F(\alpha) = \mathcal{L}_{val}(\omega^*(\alpha), \alpha)$$
$$s.t. \ \omega^*(\alpha) = \arg\min_{\omega \in \Omega} \mathcal{L}_{train}(\omega, \alpha) \quad (1)$$

where $\alpha$ is the network architecture parameters, and $\omega$ are the weights of the network. The lower-level objective $\mathcal{L}_{train}(\omega, \alpha)$ is the cross-entropy loss function on the training data. $\omega^*(\alpha)$ is the weights of the network where $\mathcal{L}_{train}(\omega, \alpha)$ is maximized. The upper-level objective $F(\alpha)$ consists of the loss function $\mathcal{L}_{val}(\omega^*(\alpha), \alpha)$ on the validation data. The goal for the bilevel architecture optimization is to find $\alpha^\#$ that minimizes the validation loss $\mathcal{L}_{val}(\omega(\alpha), \alpha)$. The topology of a block is searched in the process of finding $\alpha^\#$, and the process of architecture search is shown in Fig.1.

The topology of a block $\Theta$ is defined as a directed acyclic graph (DAG) with $N$ nodes. $\{\Theta_i, i \leq L\}$ represents one of the blocks in the network, and $\{X_0, X_1, ..., X_{N-1}\}$ represents $N$ nodes. We denote the operation space as $\mathcal{O}$ and each of the candidate operations is denoted as $o(\cdot)$. Edge $E_{i,j}$ represents the feature map connecting node $X_i$ and node $X_j$. Every edge $E_{i,j}$ consists of a set of candidate operations and the operation weights as architecture parameters $\alpha^{(i,j)}$. Therefore, edge $E_{i,j}$ can be formulated as:

$$f_{i,j}(X_i) = \sum_{o \in \mathcal{O}_{i,j}} \frac{\exp(\alpha_o^{(i,j)})}{\sum_{o' \in \mathcal{O}} \exp(\alpha_{o'}^{(i,j)})} o(X_i) \quad (2)$$

Where $i < j$. So the intermediate node between the input node and output node can be represented as $X_j = \sum_{i<j} f_{i,j}(X_i)$. Correspondingly, $X_0$ and $X_{N-1}$ represent the input node and output node, respectively. The output node is $X_{N-1} = \sum_{i=1}^{N-2} X_i$. It can be obtained from Fig. 1 (b) that the topology of the block consists of an input node, an output node, and two intermediate nodes. When designing the topology of the block, it is found through experiments that the topology of the proxy network in [12] is easy to cause gradient explosion in the proxy network training process. To adapt to the feature map in Conv operations and MHSA operations, the topology of the proxy network is shown in Fig. 1 (b). Consequently, the block $\Theta$ consists of the nodes $\{X_0, X_1, X_2, X_3\}$. Then the topology of the block can be formulated as:

$$X_1 = f_{0,1}(X_0)$$
$$X_2 = f_{0,2}X_0 + f_{1,2}(X_1) \quad (3)$$
$$X_3 = X_1 + X_2 = f_{0,2}X_0 + f_{1,2}(f_{0,1}(X_0))$$

Since the network with large depth requires a large number of GPU memory, some architecture search algorithms are searched on shallow networks and the network architecture performance is verified on deep networks. However, the performance of shallow and deep networks is very different, which usually means that the architecture chosen from the search is not necessarily the optimal evaluation [35-37]. This is known as the depth gap between search and evaluation. P-DARTS [12] finds that during multiple searches in DARTS, the block structure of the searched architecture tends to remain shallowly connected rather than deeply connected. This is due to the fact that shallow networks usually have a faster gradient descent during the search, which contradicts the attempt that deeper networks tend to obtain better performance [3, 15, 38]. However, the network depth of the search directly follows the configuration of the evaluation network, which can cause difficulties in terms of GPU memory. In order to achieve a more efficient search with limited GPU memory, a progressive search approach is applied, and the network depth is gradually increased during the search so that the depth of the network at the end of the search is close enough to the settings applied for evaluation and validation. Also, the progressive search approach is used to reduce the search space by searching in shallow networks relative to candidate operations, thus mitigating the risk of searching in deep networks. Limited GPU memory is an important issue to consider during computation, and one solution is to reduce the number of channels in each operation. To solve this problem, the number of candidate operations can be gradually reduced at the end of each stage, and the approach utilizes the architectural parameter $\alpha$ of the candidate operation as the selection criterion.

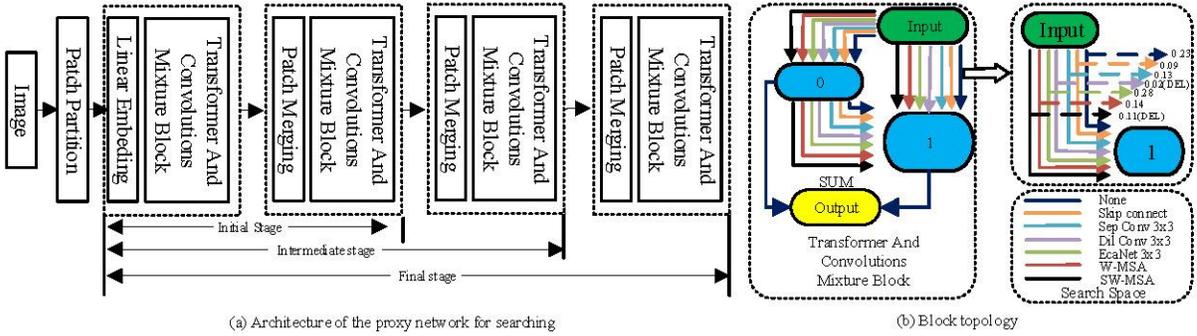

Fig.1. The overall proxy network architecture of VTCAS. The architecture search process consists of three stages: the initial stage, the intermediate stage, and the final stage. The depth of the proxy network increases from 3 at the initial stage to 4 and 5 at the intermediate stage and the final stage, while the number of candidate operations (viewed in color) between nodes is trimmed by 2, 2, and 2 respectively. Each operation between nodes is assigned a score, and the ones with the lowest-scored are dropped at the current stage. There is only one operation between each node in the final result. For brevity, only one key block of the initial stage is shown to illustrate the change of topology.

Fig.1 (a) shows the progressive architecture search approach composed of three stages: the initial stage, the intermediate stage, and the final stage. The proxy network for searching architecture is trained for 50 epochs in each stage, as suggested in [12]. In one epoch, the network weight $\omega$ and the architecture parameter $\alpha$ are updated through backpropagation separately in order, and the architecture parameter $\alpha$ has only been updated since the 21st epoch. At the beginning of the initial stage, 7 candidate operations construct edges $E_{i,j}$ between nodes, the initial dimension is 24 and there are two layers of blocks to construct a proxy network for searching the topology of blocks through VTCAS, which is shown in Fig. 1 (b). When the proxy network in the initial stage completes 50 iterations, the architecture parameters $\alpha$ are updated, and two candidate operations with the smallest architecture parameter $\alpha$ are removed from $\mathcal{O}$. There are 5 candidate operations chosen for the intermediate stage. In the intermediate stage, there are 4 candidate operations constituting each edge, the initialization dimension is 48, and there are 3 layers of blocks constituting the proxy network. As the training of the intermediate stage is completed, 2 candidate operations with the smallest architectural parameter $\alpha$ are removed from $\mathcal{O}$. There are 3 candidate operations left into the final stage. In the final stage, each edge will consist of 3 candidate operations with an initialization dimensionality 96 and a four-layer block constituting the proxy network, which is the same as the number of layers of the target network used for evaluation. Finally, we remove the 2 candidate operations with the smallest architectural parameters $\alpha$ from $\mathcal{O}$ for forming the topology of the final block.

### B. Search Space

In this work, neural architecture is to design the optimal network architecture from the search space with Conv operations and MHSA operations. The versatility of the search space has a major impact on the quality of the optimal architecture. In the modern classic network, the architecture is mostly composed of block-level design. Block-level design refers to hierarchical connection methods and calculation operations. Many manually-designed networks are based on this idea, such as ResNet [4] and Inception [38]. Similarly, various NAS methods also carry out a search for the best blocks. A block is a small module in the network that is repeated multiple times to form the entire neural network. Therefore, designing the search space should be oriented to block-level design. In this work, we leverage the partial search space of P-DARTS [12] and the MHSA operations of Swin Transformer [19] as the baseline framework.

We search architecture on the proxy network with fewer blocks. The goal of designing the search space is to obtain the blocks with Conv operations and MHSA operations. For the candidate Conv operations, we choose the computational operations between nodes from the following options, which are collected from convolution neural network and MHSA modules with excellent performance [19].
- no connect (None)
- Identity (skip-connect))
- Depthwise-separable Convolution (sep_Conv_3x3)
- Dilated Convolution (dil_Conv_3x3)
- Efficient channel attention (EcaNet_3x3)
- Window multi-head self-attention (W-MSA)
- Shifted window multi-head self-attention (SW-MSA)

As shown in Fig. 1 (b), during the search initialization at the first stage, the connection between two nodes is the operations of the search space described above, and there are 7 operations between two nodes. By the proposed approach, the final block contains only one computational operation per two nodes.

### C. Evaluation Network Architecture

Through the proposed approach, we develop the block topology, which is the basis for constructing the network. The blocks are utilized to construct a target network for verifying the performance. The pipeline of the target network is summarized in Table I. The overall pipeline of the target network still follows the work of Swin Transformer [19], and some adaptive modifications have been made. First, the input RGB image is divided into non-overlapping patches through the linear embedding layer, which is the same as the processing method in ViT [8]. In this way, each patch is denoted as an important "token" as the NLP task. The feature of the image is regarded as a two-bit matrix transformation to a one-dimensional vector, that is, the $224 \times 224 \times 3$ tensor of the image is projected to an arbitrary dimension through the linear embedding layer. Therefore, the feature map can be expressed as $X_{(B,L,4 \times C)}$, where $X$ denotes the input matrix and $(B, L, 4 \times C)$ denotes the dimensionality of the matrix $X$. A detailed introduction to $X_{(B,L,4 \times C)}$ is presented in Section III-D.

TABLE I: ARCHITECTURAL PARAMETERS OF THE NETWORK UTILIZED TO EVALUATE PERFORMANCE.

| Stage | Output Size | Name | Operation Size |
|---|---|---|---|
| | 224 * 224 | Images | channels 3 |
| | 56 * 56 | Linear Embedding | dim 96 |
| 1 | 56 * 56 | VTCAS Block | [dim 96 head 3] ×2 |
| 2 | 28 * 28 | Patch Merging + VTCAS Block | [dim 192 head 6] ×2 |
| 3 | 14 * 14 | Patch Merging + VTCAS Block | [dim 384 head 12] ×6 |
| 4 | 7 * 7 | Patch Merging + VTCAS Block | [dim 786 head 24] ×2 |
| | 1 * 1 | Linear Layer | dim 96 |

The searched blocks containing MHSA operations and Conv operations are applied to the patches composed of these tokens. The whole targets network is divided into four sections starting from the linear embedding layer. Except for the first section, the number of tokens is reduced to 25% through the patch merging layer before the blocks are applied to the patches, while the dimensionality of the blocks is increased to two times. As the depth of the network increases, both the dimensionality of the blocks and the number of heads in MHSA operation increase. The details of architecture parameters are shown in Table I. Therefore, the target network based on the block searched can output multiple numbers of tokens, corresponding to the multi-scale feature maps that can be outputted. As a result, the target network can provide a convenient alternative backbone network in existing methods on various vision tasks.

### D. Transform in Computation Operation

In training the target network used to verify the block architecture performance, we also encountered the problem of gradient disappearance. Through a detailed analysis, it has been found that the change of data dimension for feature map through Conv operations and MHSA operations cause the discontinuity of the gradient in the backpropagation process. In an address this problem, we process the change near the feature map through computation operation by making the gradient continuous by the function $h$. Here we illustrate the transformation process by the dimensional change of the feature map in the block of the first layer of the network. As in ViT [8], an input RGB image is split through the patch splitting module into non-overlapping patches. Each patch is tagged as a token and its features are set as the stitching of RGB values of the original pixels. We define RGB image as $X_{(B,H,W,C)}$ ($B$ denotes the batch size, $H$ denotes the height of the image, $W$ denotes the width of the image, and $C$ denotes the number of channels of the image). $X_{(B,H,W,C)}$ is transformed to $X_{(B,L,4 \times C)}$ by a patch partition layer and linear embedding layer, where $L = H \times W$. That is, the feature map through the block is $X_{(B,L,C)}$. Then, $X_{(B,L,C)}$ presented in the Conv operation can be formulated as:

$$\begin{aligned} X_{(B,4 \times C,H,W)}^{m_0} &= g\big(\big(h(X_{(B,L,4 \times C)}^{l-1})\big)_{(B,4 \times C,L)}\big) \\ X_{(B,4 \times C,H,W)}^{m_1} &= o(X_{(B,4 \times C,H,W)}^{m_0}) \\ X_{(B,L,4 \times C)}^{l} &= h\big(\big(g^{-1}(X_{(B,4 \times C,H,W)}^{m_1})\big)_{(B,4 \times C,L)}\big) \end{aligned} \qquad (4)$$

where $X_{(B,L,4 \times C)}^{l-1}$ represents the feature map before computation operation and $X_{(B,L,4 \times C)}^{l}$ represents the feature map through computation operation. While $h(\cdot)$ represents the transpose of tensors. $g(\cdot)$ and $g(\cdot)^{-1}$ represents the process of increasing and decreasing dimensions of tensors, respectively. Similarly, in the MHSA operation, the mapping process of $X_{(B,L,C)}$ can be formulated as:

$$X^{m_0}_{(B,H,W,4\times C)} = g\big(LN(X^{l-1}_{(B,L,4\times C)})_{B,L,4\times C}\big)$$
$$X^{m_1}_{(B,H,W,4\times C)} = o(X^{m_0}_{(B,H,W,4\times C)}) + X^{m_0}_{(B,H,W,4\times C)}$$
$$X^{m_2}_{(B,H,W,4\times C)} = \big(MLP(LN(X^{m_1}_{(B,H,W,4\times C)}))\big) \quad (5)$$
$$\qquad\qquad + X^{m_1}_{(B,H,W,4\times C)}$$
$$X^{l}_{(B,L,4\times C)} = g^{-1}(X^{m_2}_{(B,H,W,4\times C)})$$

Because the transformation of the feature map in Conv operations and MHSA operations is different, the key to ensuring that the network can be trained normally is the transformation of the feature map in the Convolutional module in (4). Here we take feature map $X^{l-1}_{(1,9,1)}$ as an example to show the mapping process of functions $h$ and $g$ for Conv operation. The Conv operation is denoted as $o_{Conv}$ (kernel size is 3, the size of the padding added for all four sides is 1, the stride of the convolution is 1, learnable bias is 3). The feature map $X^{l-1}_{(1,9,1)}$, convolution weight is $W$, and the convolution layer is $l$, and loss $L^l$ are expressed as:

$$X^{l-1}_{(1,9,1)} = \big[[[x^{l-1}_{(1,1)}],[x^{l-1}_{(1,2)}],...,[x^{l-1}_{(3,3)}]]\big]$$
$$W_{(3,3)} = \begin{bmatrix} w_{1,1} & w_{1,2} & w_{1,3} \\ w_{2,1} & w_{2,2} & w_{2,3} \\ w_{3,1} & w_{3,2} & w_{3,3} \end{bmatrix} \quad (6)$$
$$L^{l}_{(1,9,1)} = \big[[[\delta^{l}_{1,1}],[\delta^{l}_{2,1}],...,[\delta^{l}_{3,3}]]\big]$$

Then $X^{l-1}_{(1,9,1)}$ is transformed as (4), which can be formulated in the flow:

$$X^{m'}_{(1,1,9)} = h(X^{l-1}_{(1,9,1)})$$
$$\qquad = [[[x^{l-1}_{(1,1)}, x^{l-1}_{(1,2)}, ..., x^{l-1}_{(3,3)}]]]$$
$$X^{m_0}_{(1,1,3,3)} = g(X^{m'}_{(1,1,9)}) \quad (7)$$
$$\qquad = \left[\left[\begin{bmatrix} x^{l-1}_{1,1} & x^{l-1}_{1,2} & x^{l-1}_{1,3} \\ x^{l-1}_{2,1} & x^{l-1}_{2,2} & x^{l-1}_{2,3} \\ x^{l-1}_{3,1} & x^{l-1}_{3,2} & x^{l-1}_{3,3} \end{bmatrix}\right]\right]$$

$$X^{m_1}_{(1,1,3,3)} = o(X^{m_0}_{(1,1,3,3)})$$
$$= \begin{bmatrix} 0 & 0 & 0 & 0 & 0 \\ 0 & x^{l-1}_{1,1} & x^{l-1}_{1,2} & x^{l-1}_{1,3} & 0 \\ 0 & x^{l-1}_{2,1} & x^{l-1}_{2,2} & x^{l-1}_{2,3} & 0 \\ 0 & x^{l-1}_{3,1} & x^{l-1}_{3,2} & x^{l-1}_{3,3} & 0 \\ 0 & 0 & 0 & 0 & 0 \end{bmatrix}$$
$$* \begin{bmatrix} w_{1,1} & w_{1,2} & w_{1,3} \\ w_{2,1} & w_{2,2} & w_{2,3} \\ w_{3,1} & w_{3,2} & w_{3,3} \end{bmatrix} \quad (8)$$
$$= \begin{bmatrix} x^{l}_{1,1} & x^{l}_{1,2} & x^{l}_{1,3} \\ x^{l}_{2,1} & x^{l}_{2,2} & x^{l}_{2,3} \\ x^{l}_{3,1} & x^{l}_{3,2} & x^{l}_{3,3} \end{bmatrix}$$
$$X^{l}_{(1,9,1)} = h(g^{-1}(X^{m_0}_{(1,1,3,3)}))$$
$$= [[[x^{l}_{(1,1)}],[x^{l}_{(1,2)}],...,[x^{l}_{(3,3)}]]]$$

Based on (6), we need to transform the loss value $L^l$ through the mapping process of functions $h$ and $g$ to ensure the continuity of the loss function in the backpropagation process. The process of the transform can be expressed by (9). Through (9) and (4), the loss value $L^{l-1}$ of the Convolution layer $l-1$ is inferred as (10).

$$L^{m'}_{(1,1,9)} = h(L^{l}_{(1,9,1)})$$
$$\qquad = [[[\delta^{l}_{(1,1)}, \delta^{l}_{(1,2)},...,\delta^{l}_{(3,3)}]]]$$
$$L^{m_0}_{(1,1,3,3)} = g(L^{m'}_{(1,1,9)}) \quad (9)$$
$$= \left[\left[\begin{bmatrix} \delta^{l}_{1,1} & \delta^{l}_{1,2} & \delta^{l}_{1,3} \\ \delta^{l}_{2,1} & \delta^{l}_{2,2} & \delta^{l}_{2,3} \\ \delta^{l}_{3,1} & \delta^{l}_{3,2} & \delta^{l}_{3,3} \end{bmatrix}\right]\right]$$

$$L^{l-1} = h\left(g^{-1}\left(\begin{bmatrix} \delta^{l-1}_{1,1} & \delta^{l-1}_{1,2} & \delta^{l-1}_{1,3} \\ \delta^{l-1}_{2,1} & \delta^{l-1}_{2,2} & \delta^{l-1}_{2,3} \\ \delta^{l-1}_{3,1} & \delta^{l-1}_{3,2} & \delta^{l-1}_{3,3} \end{bmatrix}\right)\right)$$
$$= h\left(g^{-1}\left(\begin{bmatrix} 0 & 0 & 0 & 0 & 0 \\ 0 & \delta^{l}_{1,1} & \delta^{l}_{1,2} & \delta^{l}_{1,3} & 0 \\ 0 & \delta^{l}_{2,1} & \delta^{l}_{2,2} & \delta^{l}_{2,3} & 0 \\ 0 & \delta^{l}_{3,1} & \delta^{l}_{3,2} & \delta^{l}_{3,3} & 0 \\ 0 & 0 & 0 & 0 & 0 \end{bmatrix}\right.\right. \quad (10)$$
$$\left.\left. * \begin{bmatrix} w_{3,3} & w_{3,2} & w_{3,1} \\ w_{2,3} & w_{2,2} & w_{2,1} \\ w_{1,3} & w_{1,2} & w_{1,1} \end{bmatrix}\right)\right)$$

In the process of network training, not only the forward propagation is continuous, but we also need to make the gradient of the weight of the network continuous. Through (10) and (6), we can infer that the gradient of the Convolution weight $W$ is as in (11), where function $k(\cdot)$ represents the padding.

$$\nabla^{l}_{W} = \begin{bmatrix} \nabla^{l}_{W_{1,1}} & \nabla^{l}_{W_{1,2}} & \nabla^{l}_{W_{1,3}} \\ \nabla^{l}_{W_{2,1}} & \nabla^{l}_{W_{2,2}} & \nabla^{l}_{W_{2,3}} \\ \nabla^{l}_{W_{3,1}} & \nabla^{l}_{W_{3,2}} & \nabla^{l}_{W_{3,3}} \end{bmatrix}$$
$$= \begin{bmatrix} 0 & 0 & 0 & 0 & 0 \\ 0 & x^{l}_{1,1} & x^{l}_{1,2} & x^{l}_{1,3} & 0 \\ 0 & x^{l}_{2,1} & x^{l}_{2,2} & x^{l}_{2,3} & 0 \\ 0 & x^{l}_{3,1} & x^{l}_{3,2} & x^{l}_{3,3} & 0 \\ 0 & 0 & 0 & 0 & 0 \end{bmatrix} \quad (11)$$
$$* \begin{bmatrix} \delta^{l}_{1,1} & \delta^{l}_{1,2} & \delta^{l}_{1,3} \\ \delta^{l}_{2,1} & \delta^{l}_{2,2} & \delta^{l}_{2,3} \\ \delta^{l}_{3,1} & \delta^{l}_{3,2} & \delta^{l}_{3,3} \end{bmatrix}$$
$$= k\big(g\big(h\big([[[x^{l-1}_{(1,1)}],[x^{l-1}_{(1,2)}],...,[x^{l-1}_{(3,3)}]]]\big)\big)\big)$$
$$* g\big(h\big([[[\delta^{l}_{1,1}],[\delta^{l}_{2,1}],...,[\delta^{l}_{3,3}]]]\big)\big)$$

## IV. EXPERIMENTS DESIGN

To evaluate the performance of the proposed architecture searching algorithm, a series of experiments have been designed on image classification and image object detection. First, the proposed architecture searching algorithm is compared with the peer competitors on image classification and object detection in Subsection IV-A. Then, the benchmark datasets are detailed in Subsection IV-B. Finally, the parameter setting of the search algorithm and the parameter setting of the verification experiment will be shown in Subsection IV-C.

### A. Peer Competitors

To show the effectiveness and efficiency of the proposed architecture search, the most state-of-the-art algorithms are chosen as the peer competitors on image classification and object detection.

First, the CNNs are shown which are designed by experts who have domain knowledge of both investigated data, including ResNet [4], VGGNet [3], Inception [36, 38], and Dual Path Networks (DPN) [39]. Specifically, two versions of ResNet Architectures will be obtained, that is the ResNet with the depth of 50 and 152, which can be expressed as ResNet-50 and ResNet-101. For Inception, there are also two versions considered as Inception-v1 [38] and Inception-v2 [36]. For DPN [39], we use the version with the depth of 98 and 131, which are simplified as DPN-98 and DPN-131.

The second part contains networks based on Transformer architectures. In particular, Swin Transformer [19], ViT [8], and DeiT [18] are chosen as the competitors. To be fair, the network architectures with similar parameter sizes are chosen to be compared separately. Specifically, Swin Transformer [19] is the main literature we follow and two versions Swin-T and Swin-B are chosen to be compared separately. Similarly, we chose two versions of the architecture network ViT-B and Vit-L for comparison of classification accuracy [8]. Likewise, we chose two architectures, DeiT-S and DeiT-B, for comparison in [18].

The third part includes the CNN architecture search algorithm with automatic design. Especially, Genetic CNN [27], Hierarchical Evolution [40], NASNet [32], Efficientnet-B5 [41], Block-QNN [34], DARTS [13], PC-DARTS [14], and P-DARTS [12] belong to mainly three pathways for NAS, that are evolutionary algorithm, reinforcement learning, and gradient-based NAS.

In addition, when comparing the proposed algorithm with competitors when applied to problems of image classification, we make detailed system-level comparisons. For the fairness of comparison, we utilize the same typical object detection framework to compare the detection performance with different backbone architectures. Typical object detection frameworks include: Faster RCNN [42], Cascade Mask R-CNN [43, 44], ATSS [45] and Deformable DETR [46] in MMDetetion [47]. For the above four vision task frameworks, the same training setting [48, 49] to adjust the size of the image input is utilized so that the short side is between 480 and 800, and the long side is at most 1333. The optimizer is chosen as AdamW [50], and the learning rate, weight decay, and schedules are set to the same parameters separately according to the dataset.

### B. Benchmark Datasets

For image classification, the proposed image classification standard network searching by VTCAS is benchmarked on ImageNet-1K [1], which contains 1.28M images and 50K validation images. ImageNet-1K consists of 1K object categories and the input image size is fixed to be $224 \times 224$. In particular, to reduce the search time, we follow the work [12] and randomly sample two subsets from the ImageNet training set as the training set and the validation set for searching network architecture, which is $10\%$ and $2.5\%$ images respectively. The Top-1 accuracy is used as the evaluation standard.

For the image object detection, the obtained experimental results are completed on three public datasets. The public datasets include COCO 2017 [51], VOC2007+2012 [52, 53], and ExDark [54]. The industrial production dataset is a chip defect detection dataset (DHU-CDD). 118K training set images and 5K validation set images are included in COCO 2017 to investigate the feature extraction capability of the backbone with VTCAS. VOC2007+2012 consists of 20 classes formed by merging the datasets PASCAL VOC 2007 and PASCAL VOC 2012, where the training set is the training-validation set (5011 images) of PASCAL VOC 2007 and the training-validation set (11540 images) of PASCAL VOC 2012, and the test set (4952 images) of PASCAL VOC 2007 forms the test set of VOC2007+2012. ExDark [54] is a dataset to study object detection in a low illumination environment consisting of 20 classes, which collects 7363 images in a low light environment, of which the training set contains 4800 images and the test set contains 2563 images.

### C. Parameter Settings

According to the above discussion, the experiment consists of three parts: architecture search, performance verification for image classification, and performance verification for object detection. In order to further improve the application ability of the proposed algorithm, some of the experimental settings are followed from P-DARTS [12], Swin-Transformer [19], and MMDetection [47], and modify parameter settings is according to the actual experimental scene. For the architecture search part, the network architecture is searched by the proposed VCTAS algorithm on the proxy network. The architecture search process includes three stages: the initial, the intermediate, and the final stage, as suggested in [12]. Based on the variation of the depth of the proxy network utilized for architectural search in [12] and [19] and the architecture of MHSA, the proxy network depth is increased from 3 in the initial stage to 4 and 5 in the intermediate and final stages, respectively. At the same time, the number of candidate operations between nodes is reduced by 2, 2, and 2, respectively, and each stage is trained for 50 epochs, which follows [12]. The parameters settings for the search process on proxy network architecture shown in Table II are referenced from the code open-sourced in [12], except for the operation size, which is set according to the search space.

TABLE II: PARAMETER SETTING IN ARCHITECTURE SEARCH PART.

| Names | Initial | Intermediate | Final |
|---|---|---|---|
| Init Channels | 24 | 72 | 96 |
| Network Depth | 2 | 3 | 4 |
| Operation Size | 7 | 5 | 3 |
| Stage Epochs | 50 | 50 | 50 |
| Start Arch. Epoch | 20 | 20 | 20 |

For the block architecture obtained from the architecture search on the image classification, we employ AdamW [50] as the optimizer, and the learning rate is scheduled using the cosine decay learning rate method with 20 epochs as a linear warm-up, as suggested in [19]. The initialized learning rate is 0.001 and the weight decay is 0.05, followed by the contained in [8]. In [19], the AdamW [50] optimizer is run 300 epochs. However, we find in our experiments that the training process of the target network architecture obtained by VTCAS search is prone to gradient explosion, so we reduced the learning rate to 0.0008 and set the number of epochs to 400. At the same time, the training strategy includes most of the enhancement and regularization strategies in [18]. For object detection, we conducted experiments on four datasets. For the experiments on COCO2017, and ExDark, the same settings were utilized for all object detection frameworks: multiscale training [48, 49] (resizing the inputs images size that the side is resized as between 480 and 800, and the longer side is at most 1333), AdamW [50] optimizer (initialized with a learning rate of 0.0001 and the weight decay of 0.05), batch size with 16 and 3x schedule (36 epochs). For the experiment on VOC, we followed the settings in MMDetection [47], SGD optimizer, with a learning rate of 0.01 and 8 epochs. In particular, all our experiments are based on Pytorch 1.9 and performed on four GPU cards with Nvidia Geforce RTX 3090.

## V. EXPERIMENTS RESULTS AND ANALYSIS

In this section, we first show the results on the network architecture searching on ImageNet subset and describe the composition of the searched architecture in detail. Then, the performance results of the target network composed of blocks from VTCAS on ImageNet are shown in detail and compared with other competitors. Finally, we will show the performance of various object detection frameworks consisting of the backbones from VTCAS on four datasets.

### A. Vision Transformer with Convolutions Block Architecture

Based on the ImageNet subset [14], the block architecture discovered by VTCAS is shown in Fig. 2. To further illustrate the topology of the obtained blocks, the feature map of the input node is $X_0$ and the feature map of the output node is $X_3$. The feature maps at node 0 and node 1 are $X_1$ and $X_2$, respectively. It can be seen from Fig. 2 that $X_0$ is operated by $3 \times 3$ separable convolutions computation operation to obtain $X_1$ on node 0 and $3 \times 3$ efficient channel attention convolutions computation operation to obtain $X_2'$ on node 1, respectively. Then, $X_1$ is computed by SW-MSA to obtain $X_2''$ and summed with the feature map of node 1 to obtain $X_2 = X_2' + X_2''$. Finally, $X_1$ of node 0 is added with $X_1$ of node 1 to get the feature map $X_3 = X_1 + X_2$ of the output node. We can simplify the whole block in the form:

$$\begin{aligned} X_1 &= o_{(Seq\_Conv\_3*3)}(X_0) \\ X_2 &= o_{(EcaNet\_3*3)}(X_0) + o_{(SW-MSA)}(X_1) \\ X_3 &= X_1 + X_2 \end{aligned} \quad (12)$$

Compared to the topology from P-DARTS [12], the block architecture searched contains only one input node, while the block architecture from P-DARTS contains two input nodes. In addition, between the input and output nodes, we have only two intermediate nodes, while the block architecture from P-DARTS contains four intermediate nodes. The proposed method searches a relatively simple topology for the purpose of the proposed algorithm is to obtain a more efficient combination of MHSA [19] and CNN to identify objects with complex structures, and a simple topology allows the proposed algorithm to obtain the optimal topology faster.

### B. Performance verification on image classification

Table III shows the comparison of the classification accuracy of the proposed algorithm with other peer competitors on ImageNet, and '-' indicates that the relevant literature does not provide the corresponding performance value. Compared to other manually CNN networks, such as ResNet-50 [4], DPN-98 [39], etc., the network composed of blocks from VTCAS is with less parameter size and higher classification accuracy. Also, the proposed algorithm partially follows the related work of Swin-Transformer [19], so Swin-T and Swin-B are chosen as comparison networks. Compared to Swin-T, VTCAS is slightly larger in terms of parameter size and floating-point operations, and the classification accuracy is higher than Swin-T by $0.7\%$. Compared with other Vision Transformer-based networks, VTCAS obtain higher classification accuracy and less parameter size. To CNN-based NAS, the classification accuracy of VTCAS is higher than most NAS methods except Efficientnet-B5 [41]. VTCAS and Efficientnet-B5 with similar parameter sizes. However, VTCAS is with an input image size of 224, and Efficientnet-B5 is 456 so the performance of Efficientnet-B5 is higher than VTCAS.

### C. Performance Analysis on Object Detection

Table IV shows the results of object detection and instance segmentation experiments on COCO 2017. The experiments are based on MMDetection [47], and Cascade Mask R-CNN [43, 44] is chosen as the basic framework of object detection. We compare VTCAS with standard CNNs (i.e., ResNet-50) and Vision Transformer networks (i.e., DeiT, Swin). The comparison is conducted only by changing the backbones and leaving the other training settings unchanged. It should be noted that due to the hierarchical feature maps of VTCAS, Swin Transformer [19], and ResNe(X)t [37], they can be directly applied to all the above frameworks. Since DeiT [18] only produces a single feature map resolution, it cannot be directly applied. For a fair comparison, the approach [55] is followed to construct hierarchical feature maps using deconvolution layers for DeiT.

According to the analysis in Table IV, compared with ResNet-50, the backbone based on VTCAS brings a box AP gain of $4.1\%$ and a mask AP gain of $3.4\%$, but the network parameter size and FLOPs are slightly higher than Resnet-50. Similarly, as the backbone of cascade mask RCNN, the performance of VTCAS is higher than DeiT-S and lower than Swin-T. The result of VTCAS are $+2.4$ box AP and $+2.3$ mask AP higher than DeiT-S but $-0.1$ box AP and $-0.1$ mask AP are less than Swin-T. In the classification task of ImageNet-1K, the VTCAS parameters are larger than Swin-T and DeiT-S, and the backbone of Cascade Mask RCNN is based on the pre-training of ImageNet-1K, so the parameter size of VTCAS-based Cascade Mask RCNN is higher than that of DeiT-S and Swin-T.

To further verify the ability for image feature extraction with the architecture searched by VTCAS, Table V lists the results of VTCAS, Swin-T, and ResNet-50 on the four kinds of object detection frameworks. The experimental results are all completed on VOC2007+2012, and these object frameworks are with the same training configuration. The experiment compares the feature extraction capabilities of a network only by changing the backbones of the object detection framework and keeping other settings unchanged.

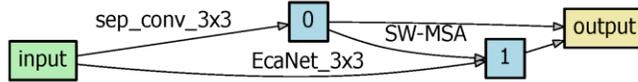

Fig.2. Block Architecture searching by VTCAS. Block contains four nodes, one input node, one output node, and two intermediate nodes (node 0 and node 1). The nodes are linked by edges, and each edge represents a computational operation. The feature map through the block from the input node and the data stream outputs the block from the output node, and feature map passes through the computation operation represented by each edge in turn.

TABLE III: COMPARISON BETWEEN THE PROPOSED ALGORITHM AND THE STATE-OF-THE-ART PEER COMPETITORS IN TERMS OF THE CLASSIFICATION ACCURACY (TOP-1) ON IMAGENET.

| Method Type | Network | Image Size | #Param.(M) | FLOPS (G) | ImageNet Top 1 (%) |
|---|---|---|---|---|---|
| manually designed CNNs | ResNet-50 [4] | 224*224 | 26.0 | - | 75.3 |
| | ResNet-152 [4] | 224*224 | 60.0 | - | 77 |
| | VGGNet [3] | 224*224 | 138.0 | - | 71.5 |
| | Inception-v1 [38] | 224*224 | 7.0 | - | 69.8 |
| | Inception-v2 [36] | 224*224 | 11.0 | - | 74.8 |
| | DPN-98 [39] | 224*224 | 61.7 | 11.7 | 79.8 |
| | DPN-131 [39] | 224*224 | 80.0 | 16.0 | 80.1 |
| manually designed Vison Transformer | Swin-T [19] | 224*224 | 29.0 | 4.5 | 81.3 |
| | Swin-B [19] | 224*224 | 88.0 | 8.7 | 83.0 |
| | Vit-B [8] | 384*384 | 86.0 | 55.4 | 77.9 |
| | Vit-L [8] | 384*384 | 307.0 | 190.7 | 76.5 |
| | DeiT-S [18] | 224*224 | 22.0 | 4.6 | 79.1 |
| | DeiT-B [18] | 224*224 | 86.0 | 17.5 | 81.8 |
| automatic | Genetic CNN [27] | 224*224 | 156.0 | - | 72.1 |
| | Hierarchical Evolution [40] | 224*224 | - | - | 79.7 |
| | NASNet-A [32] | 224*224 | 5.3 | 0.6 | 74.0 |
| | NASNet-B [32] | 224*224 | 5.3 | 0.5 | 72.8 |
| | Efficientnet-B5 [41] | 456*456 | 30.0 | 9.9 | 83.6 |
| | Block-QNN-S [34] | 224*224 | 95.0 | - | 78.1 |
| | DARTS [34] | 224*224 | 4.7 | - | 73.3 |
| | PC-DARTS [14] | 224*224 | 5.3 | - | 73.8 |
| | P-DARTS [12] | 224*224 | 4.9 | - | 73.6 |
| | VTCAS (ours) | 224*224 | 31.2 | 5.2 | 82.0 |

TABLE IV: VARIOUS BACKBONE W. CASCADE MASK R-CNN ON COCO 2017.

| | $AP^{box}$ | $AP_{50}^{box}$ | $AP_{75}^{box}$ | $AP^{maks}$ | $AP_{50}^{maks}$ | $AP_{75}^{maks}$ | #Param(M) | FLOPs(G) |
|---|---|---|---|---|---|---|---|---|
| ResNet50 | 46.3 | 64.3 | 50.3 | 40.1 | 61.7 | 43.4 | 82.0 | 739.0 |
| DeiT-S | 48.0 | 67.2 | 51.7 | 41.4 | 64.2 | 44.3 | 80.0 | 889.0 |
| Swin-T | 50.5 | 69.3 | 54.9 | 43.7 | 66.6 | 47.1 | 86.0 | 745.0 |
| VTCAS | 50.4 | 68.9 | 54.8 | 43.6 | 66.5 | 47.1 | 90.0 | 757.6 |

TABLE V: RESULTS ON VOC2007+2012 OBJECT DETECTION.

| Detection Framework | Backbone | mAP(%) | FLOPs(G) | #Param(M) |
|---|---|---|---|---|
| Faster RCNN | ResNet50 | 80.8 | 206.76 | 41.22 |
| | SWIM | 82.4 | 210.4 | 44.84 |
| | VTCAS | 83.5 | 225.64 | 49.3 |
| Cascade RCNN | ResNet50 | 80.4 | 234.52 | 68.98 |
| | SWIM | 82.9 | 238.17 | 72.61 |
| | VTCAS | 83.1 | 253.4 | 77.07 |
| ATSS | ResNet50 | 78.1 | 202.35 | 31.93 |
| | SWIM-T | 81.1 | 208.61 | 35.6 |
| | VTCAS-T | 82.4 | 223.85 | 40.06 |
| Deformable Detr | ResNet50 | 73.9 | 199.88 | 40.83 |
| | SWIM-T | 79.1 | 205.37 | 41.6 |
| | VTCAS-T | 80.3 | 220.61 | 46.05 |

As can be seen in Table V that in the frameworks based on Faster-CNN, Cascade RCNN, ATSS, and Deformable Detr, VTCAS is $2.7\%$, $2.7\%$, $4.3\%$ and $6.4\%$ box mAP(0.5 IOU) higher than ResNet-50. Similarly, VCTAS outperforms Swin-T by $1.1\%$, $0.2\%$, $1.3\%$, and $1.2\%$ box AP(0.5 IOU) among the four object detection frameworks, respectively. However, the VTCAS-based object detection network is with a larger number of parameters and network complexity. According to the above analysis, VTCAS not only obtains better performance in classification tasks but also improves accuracy as a backbone for object detection.

To further illustrate the robustness of the network constructed by VTCAS on the object detection task, we experiment with the performance of the Cascade Mask R-CNN network from COCO2017 in a real-world scene. Fig. 3 shows the object detection results of the Cascade Mask R-CNN network based on ResNet-50, Swin-T, and VTCAS under low illumination and bright conditions in the same indoor scene. Fig. 3 (a)-(d) show the images in bright conditions and the most frequent objects in the images are books. Comparing Fig. 3 (d) with Figs. 3 (b)-(c), it can be seen that Fig. 3 (d) recognizes the most books. In particular, some locations subjected to specular reflections are not recognized by Figs. 3 (b)-(c) for books, and Fig. 3 (d) recognizes them. Similarly, Fig. 3 (e)-(h) are the images of the interior in low illumination indoor scene, which show the same items as Fig. 3 (a)-(d). Obviously, the books recognized in Fig. 3 (h) are the most numerous. Both Fig. 3 (d) and Fig. 3 (h) are the results of VTCAS-based Cascade Mask R-CNN network recognition. Based on the above experimental results, we can analyze that the VTCAS-based network obtains better feature extraction ability in low illumination indoor scenes. To further verify the feature extraction ability of VTCAS in low illumination indoor scenes, we conducted ExDark-based comparison experiments as in Table VI.

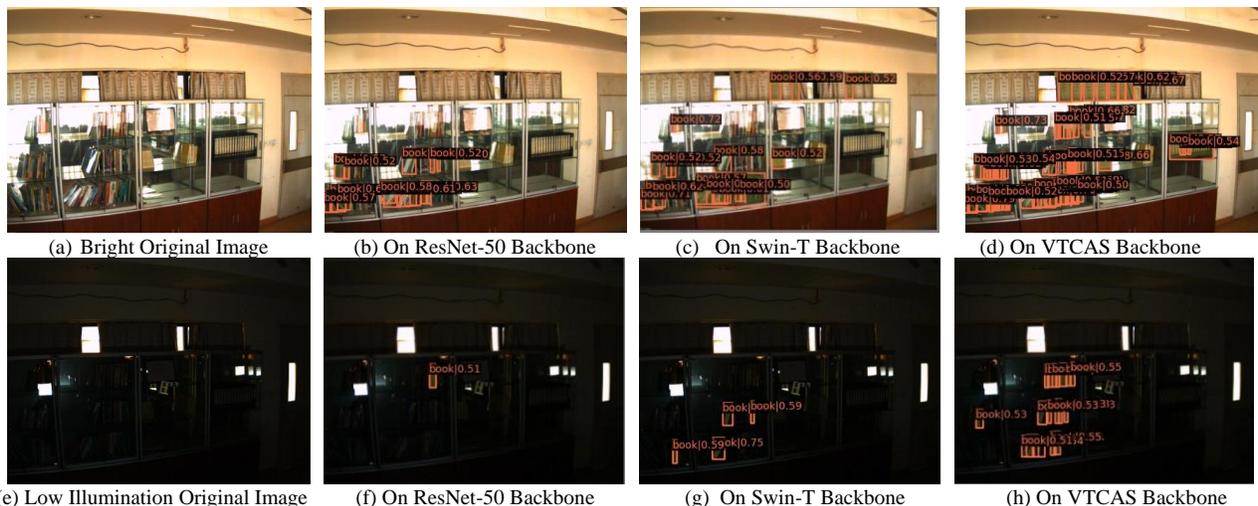

Fig.3. Visualization of the detection performance on three kinds of backbone (ResNet-50, Swin-T, and VTCAS) Cascade Mask R-CNN networks with the weight trained on COCO2017. All test images were shot in the same indoor scene, only the illumination conditions are changed. (a)The bright original image. (b) The detection result of ResNet-50-based Cascade Mask R-CNN networks in the bright indoor scene. (c) The detection result of Swin-T-based Cascade Mask R-CNN networks in the bright indoor scene. (d) The detection result of VTCAS-based Cascade Mask R-CNN networks in the bright indoor scene. (e) The low illumination original image. (f) The detection result of ResNet-50-based Cascade Mask R-CNN networks in the low illumination indoor scene. (g) The detection result of Swin-T-based Cascade Mask R-CNN networks in the low illumination indoor scene. (h) The detection result of VTCAS-based Cascade Mask R-CNN networks in the low illumination indoor scene.

TABLE VI: VARIOUS BACKBONE WITH CASCADE MASK R-CNN ON EXDARK [54].

| Class | ResNet50 | | Swin-T | | VTCAS | |
|---|---|---|---|---|---|---|
| | Recall | AP | Recall | AP | Recall | AP |
| Bicycle | 70.3 | 65.3 | 78.5 | 75.0 | 76.6 | 73.9 |
| Boat | 60.8 | 55.0 | 73.6 | 68.5 | **76.1** | **69.7** |
| Bottle | 58.7 | 55.6 | 67.0 | 64.0 | **71.6** | **67.3** |
| Bus | 81.1 | 78.0 | 87.8 | 85.6 | 86.6 | 84.2 |
| Car | 64.4 | 60.5 | 74.4 | 69.7 | 74.2 | 69.5 |
| Cat | 68.9 | 57.0 | 77.6 | 70.2 | **77.9** | **71.5** |
| Chair | 51.4 | 45.8 | 60.4 | 53.5 | 60.9 | 53.4 |
| Cup | 59.6 | 53.8 | 65.7 | 59.6 | **69.1** | **62.3** |
| Dog | 73.7 | 66.4 | 81.8 | 78.4 | **82.0** | **79.2** |
| Motorbike | 59.5 | 51.5 | 65.7 | 58.1 | **69.0** | **62.1** |
| People | 66.2 | 61.6 | 72.4 | 67.5 | **73.5** | **68.7** |
| Table | 42.8 | 27.2 | 54.3 | 38.2 | 54.3 | **38.6** |
| Average | 63.1 | 56.5 | 71.6 | 65.7 | **72.7** | **66.7** |

Table VI shows the experiments performed on ExDark [54], whose data were preprocessed according to the format of the VOC2007+2012, so AP was calculated as $IOU = 0.5$. The results in Table VI are obtained on the Cascade Mask R-CNN network with three kinds of backbone (ResNet-50, Swin-T, and VTCAS). From the analysis in Table VI, it can be obtained that the mAP of the network with VTCAS is $10.2\%$ higher than that with ResNet50 and higher than that of the network with Swin-T by $1.1\%$. Correspondingly, in the recall rate, the VTCAS-based network is $9.5\%$ higher than the ResNet50-based network and $1.1\%$ higher than the Swin-T-based network. Based on the above analysis, we can conclude that VTCAS shows higher robustness for object detection in low illumination scenes. The object detection network based on VTCAS can detect more objects in the low illumination environment. We speculate that the excellent performance of VTCAS in low illumination scenes may be related to the topology of the block as Fig. 2. In this block, the Conv operation is applied in feature map for noise reduction and multi-scale feature extraction (reducing the effect caused by low illumination on the object) at first, followed by the MHSA operation to extract the necessary features for complex objects, which may be the reason for the strong recognition capability of VTCAS in low illumination scenes.

## VI. CONCLUSIONS AND FUTURE WORK

In this work, to address the problem that ViT is not deeply optimized for object scaling and image noise in vision tasks, we propose an architectural search method, called Visual Transformer and Convolutional Architecture Search (VTCAS), that merges the benefits of MHSA with the merits of CNNs for image recognition. The proposed method effectively combines the complex pixel relationship extraction from MHSA and the scaling invariance for local objects from convolution. The search block-based backbone network can extract feature maps at different scales, which makes it compatible with a wider range of vision tasks, such as image classification and object detection. In particular, VTCAS has shown robustness and superior performance in low-illumination object detection scenarios. The existing experimental structure shows that our VTCAS architecture achieves good results and is more suitable for scenarios with complex features than the traditional Transformer architecture and CNN architecture. In the future, we will further distill the network structure and adapt VTCAS to the lightweight network to improve the recognition speed.